\documentclass{article}

 \usepackage[preprint]{neurips_2018}

\usepackage[utf8]{inputenc} % allow utf-8 input
\usepackage[T1]{fontenc}    % use 8-bit T1 fonts
\usepackage{hyperref}       % hyperlinks
\usepackage{url}            % simple URL typesetting
\usepackage{booktabs}       % professional-quality tables
\usepackage{amsfonts}       % blackboard math symbols
\usepackage{nicefrac}       % compact symbols for 1/2, etc.
\usepackage{microtype}      % microtypography

\usepackage{amsmath}
\usepackage{times}
\usepackage{graphicx}
\usepackage{color}
\usepackage{multirow}
\usepackage{setspace}
\usepackage{rotating}
\usepackage{bbm}
\usepackage{latexsym}
\usepackage[english]{babel}

\usepackage{subfig}
\usepackage{caption}
\usepackage{amssymb}
\usepackage{epsfig}
\usepackage{mathabx}

\title{Conditionally-additive-noise Models for Structure Learning}

\author{%
  Daniel Chicharro\\
  Neural Computation Laboratory\\
  Center for Neuroscience and Cognitive Systems@UniTn\\
  Istituto Italiano di Tecnologia\\
  38068 Rovereto, Italy\\
  Department of Neurobiology\\
  Harvard Medical School\\
  Boston, MA 02115\\
  \texttt{daniel.chicharro@iit.it} \\
  \texttt{daniel\_chicharro@hms.harvard.edu} \\
  \And
   Stefano Panzeri\\
   Neural Computation Laboratory\\
   Center for Neuroscience and Cognitive Systems@UniTn\\
   Istituto Italiano di Tecnologia\\
   38068 Rovereto, Italy\\
   \texttt{stefano.panzeri@iit.it} \\
   \And
   Ilya Shpitser\\
   Department of Computer Science\\
   Whiting School of Engineering\\
   Johns Hopkins University\\
   \texttt{ilyas@cs.jhu.edu} \\
}

\begin{document}

\maketitle

\begin{abstract}
Constraint-based structure learning algorithms infer the causal structure of multivariate systems from observational data by determining an equivalent class of causal structures compatible with the conditional independencies in the data. Methods based on additive-noise (AN) models have been proposed to further discriminate between causal structures that are equivalent in terms of conditional independencies. These methods rely on a particular form of the generative functional equations, with an additive noise structure, which allows inferring the directionality of causation by testing the independence between the residuals of a nonlinear regression and the predictors (nrr-independencies). Full causal structure identifiability has been proven for systems that contain only additive-noise equations and have no hidden variables. We extend the AN framework in several ways. We introduce alternative regression-free tests of independence based on conditional variances (cv-independencies). We consider conditionally-additive-noise (CAN) models, in which the equations may have the AN form only after conditioning. We exploit asymmetries in nrr-independencies or cv-independencies resulting from the CAN form to derive a criterion that infers the causal relation between a pair of variables in a multivariate system without any assumption about the form of the equations or the presence of hidden variables.
\end{abstract}

\section{Introduction}
\label{s1}

Inferring the causal structure of multivariate systems from observational data has become an indispensable need in many domains of science, from physics, to neuroscience, to finance \citep{Lutkepohl06, Wibral14, Petersbook}. Constraint-based structure learning algorithms have been used to infer the causal structure by determining an equivalent class of causal structures compatible with the conditional independencies in the data \citep{Spirtes00, Pearl09}. Additive-noise (AN) models were proposed as powerful solutions that allow further discriminating between structures within these equivalent classes \citep{Hoyer2009, Peters2014, Mooij16}. A pure AN functional equation requires that the noise is additively separable from the causes of a variable, and in the standard approach this property is exploited by testing independencies of the residuals of a nonlinear regression with the regression predictors. For multivariate systems, algorithms testing these nonlinear regression residuals independencies (nrr-independencies) proceed inferring a global causal ordering \citep{Mooij09} under the assumption that the noise is separable in all equations. This approach has been mostly studied in the case of causal sufficiency (no hidden variables), \citep[but see][for an exception]{Janzing09b}. In this work we extend the AN framework in four fronts. First, allowing for the presence of hidden variables. Second, considering functional equations that have the AN form only after conditioning on certain variables. Third, introducing an alternative regression-free test to infer causality exploiting the independencies present in AN models. Fourth, proposing a criterion to infer the causal relation between a specific pair of variables in a multivariate system with hidden variables, without restrictions on the form of the functional equations and without involving the inference of a global causal ordering.

In more detail, we generalize AN models to partial conditionally-additive-noise (CAN) models with hidden variables. These models contain both equations reducible and irreducible to the AN form, and the AN form may only be obtained after conditioning on some of the observable variables. We show how structure learning for partial CAN models can be formulated in terms of nrr-independencies asymmetries, analogously to AN models \citep{Hoyer2009, Peters2014}. Furthermore, we introduce a regression-free test to detect additive noise. This test assesses the independence of the residuals second-order moments from the predictors indirectly, estimating conditional variance independencies (cv-independencies) that do not require an actual reconstruction of the noise variables. We formulate a criterion to infer a potential cause from one particular variable to another in the presence of hidden variables, which does not require inferring a global causal ordering. Finally, we discuss the extension of CAN models by generalizing post-nonlinear AN models \citep{Zhang09}, which allow for the presence in the functional equations of a global invertible nonlinear transformation of the AN terms. We believe that this work will lead to a structure learning algorithm alternative to the ones existing for AN models with no hidden variables \citep{Mooij09, Peters2014, Buhlmann14}. The proposal of such algorithm exploiting the new criterion we propose is left for a future contribution.

This paper is organized as follows. In Section \ref{s2}, we review previous work on AN models and post-nonlinear AN models. In Section \ref{s3}, we describe the regression-free test based on cv-independence. In Section \ref{s4}, we extend the AN models to CAN models, providing conditions for the existence of cv-independencies and nrr-independencies that appear after conditioning. We introduce a criterion that exploits these independencies to infer causal relations in the presence of hidden variables and for system which may be only partially CAN models. In Section \ref{s5}, we examine examples of concrete systems. In Section \ref{s6}, we extend our approach to post-nonlinear AN models.

\section{Previous work on additive-noise models}
\label{s2}

We start with some basic notions for graphs. We use capital letters for random variables and bold letters for sets and vectors. Consider a set of random variables $\mathbf{V} = \{ \mathrm{V}_1,...,\mathrm{V}_n\}$. A graph $\mathcal{G} = (\mathbf{V}; \mathcal{E})$ consists of nodes $\mathbf{V}$ and edges $\mathcal{E}$  between the nodes. $(\mathrm{V}; \mathrm{V}) \notin \mathcal{E}$ for any $\mathrm{V} \in \mathbf{V}$. We write $\mathrm{V}_i \rightarrow \mathrm{V}_j$ for $(\mathrm{V}_i;\mathrm{V}_j)\in \mathcal{E}$. We refer to $\mathrm{V}$ as both variable $\mathrm{V}$ and its corresponding node. A node $\mathrm{V}_i$ is called a parent of $\mathrm{V}_j$ if $(\mathrm{V}_i;\mathrm{V}_j)\in \mathcal{E}$. The set of parents of $\mathrm{V}_j$ is denoted by $\mathbf{Pa}_j$. A path in $\mathcal{G}$ is a sequence of (at least two) distinct nodes $\mathrm{V}_1, ... , \mathrm{V}_n,$ such that there is an edge between $\mathrm{V}_k$ and $\mathrm{V}_{k+1}$ for all $k = 1, ... , n-1$. If all edges are $\mathrm{V}_k \rightarrow \mathrm{V}_{k+1}$ the path is a causal or directed path. A node $\mathrm{V}_i$ is a collider in a path if it has incoming arrows $\mathrm{V}_{i-1} \rightarrow \mathrm{V}_i \leftarrow \mathrm{V}_{i+1}$ and is a noncollider otherwise. The set of descendants $\mathbf{D}(\mathrm{V}_i)$ of node $\mathrm{V}_i$ comprises those variables that can be reached going forward through causal pathways from $\mathrm{V}_i$. The set of non-descendants $\mathbf{ND}(\mathrm{V}_i)$ of $\mathrm{V}_i$, is complementary to $\mathbf{D}(\mathrm{V}_i)$, including $\mathrm{V}_i$. In Directed Acyclic Graphs (DAGs) no node is its own descendant. Two nodes $\mathrm{V}_i$ and $\mathrm{V}_j$ are adjacent if either $(\mathrm{V}_i; \mathrm{V}_j) \in \mathcal{E}$,  $(\mathrm{V}_j; \mathrm{V}_i) \in \mathcal{E}$, or there is a hidden common parent $\mathrm{V}_k$ between them (i.\,e.\,$(\mathrm{V}_k; \mathrm{V}_i) \in \mathcal{E}$ and $(\mathrm{V}_k; \mathrm{V}_j) \in \mathcal{E}$ and $\mathrm{V}_k$ is not observable). $\mathrm{V}_i$ is a potential cause of $\mathrm{V}_j$ if it is a parent or they share a hidden common parent. Conditional independence between two variables is equivalent to d-separation \cite{Pearl09} of their corresponding nodes under the faithfulness assumption \cite{Spirtes00}, which ensures that the probability distribution contains only independencies induced by the causal structure. Accordingly, a conditional dependence between $\mathrm{V}_1$ and $\mathrm{V}_n$ given $\mathbf{S}$, i.e.\,, $\mathrm{V}_1 \notperp \mathrm{V}_n|\mathbf{S}$ exists iff the nodes are connected by a path that is active when blocking the nodes in $\mathbf{S}$ (S-active path). See \cite{Spirtes00} for a more detailed description.

The functional equation generating variable $\mathrm{V}_i$ has the AN form if it conforms to
\begin{equation}
\label{e0}
\mathrm{V}_i = f_i(\mathbf{V}_i,\varepsilon_i) = f_{i}(\mathrm{V}_{i,1},...,\mathrm{V}_{i,n})+ \varepsilon_i,
\end{equation}
with $\mathbf{\mathbf{Pa}}_i = \mathbf{V}_i = \{ \mathrm{V}_{i,1},...,\mathrm{V}_{i,n} \}$ and noise $\varepsilon_i$ by definition independent of the parents. Most part of the work with AN models assumes that all variables are observable. Under the assumption of no hidden variables, since the noise is additively separable from the parents, an estimate $\hat{\varepsilon}_i$ can be obtained by nonlinear regression as $\hat{\varepsilon}_i \equiv \mathrm{V}_i-\hat{f}_i(\mathbf{V}_i)$. If $\varepsilon_i$ is properly reconstructed, $\hat{\varepsilon}_i \perp \mathrm{V}\ \forall \mathrm{V} \in \mathbf{V}_i$, that is, the independence of the noise from the parents is recovered. Consider a particular variable $\mathrm{Y}$ and parent $\mathrm{X} \in \mathbf{Pa}_y$. If all variables are observed and the equation of $\mathrm{Y}$ has the AN form it is guaranteed that an independent noise can be reconstructed.

\vspace*{1mm}
\noindent \textbf{Proposition $\mathbf{1}$}\ \ \emph{Nrr-independence with AN functional equations:} `If the functional equation of $\mathrm{Y}$ has the AN form, then  $ \forall \mathrm{X} \in \mathbf{\mathbf{Pa}}_y \ \exists \mathbf{S}$ and $\hat{f}_y(\mathrm{X},\mathbf{S})$ such that $\hat{\varepsilon}_y \perp \mathrm{X}$, with $\hat{\varepsilon}_y \equiv \mathrm{Y}-\hat{f}_y(\mathrm{X},\mathbf{S})$.'

\vspace*{1mm}
The existence of at least one set $\mathbf{S}$ is guaranteed because $\mathbf{S} = \mathbf{\mathbf{Pa}}_y \backslash \mathrm{X}$ leads to $\hat{\varepsilon}_y \perp \mathrm{X}$. If we knew that $\hat{\varepsilon}_y$ reconstructs a truly generative noise variable, Proposition $1$ would suffice to infer a cause from $\mathrm{X}$ to $\mathrm{Y}$ (assuming no hidden variables). This is because, if $\mathrm{Y}$ is adjacent to both $\mathrm{X}$ and $\hat{\varepsilon}_y$ (there are edges $\mathrm{Y} \-- \mathrm{X}$ and $\mathrm{Y} \-- \hat{\varepsilon}_y$), the fact that $\mathrm{X} \notperp \mathrm{Y}$ and $\mathrm{X} \perp \hat{\varepsilon}_y$ is a sufficient condition for $\mathrm{Y}$ to be a collider ($\mathrm{X} \rightarrow \mathrm{Y} \leftarrow \hat{\varepsilon}_y$) \citep{Pearl09}. However, because $\hat{\varepsilon}_y$ is only a reconstruction of the presumed underlying noise variable, extra checks are required: the question is if the nrr-independence $\mathrm{X} \perp \hat{\varepsilon}_y$ could also occur when $\hat{\varepsilon}_y$ is estimated but the generative model contains the reverse causal relation.

\cite{Hoyer2009} proved that, if $\mathrm{Y}$ has a generative AN functional equation and $\mathrm{X} \in \mathbf{\mathbf{Pa}}_y$, nrr-independence holds for $\hat{\varepsilon}_y = \mathrm{Y}-\hat{f}_y(\mathrm{X},\mathbf{S})$, given $\mathbf{S} = \mathbf{\mathbf{Pa}}_y \backslash \mathrm{X}$ fixed, and in general not for the direction opposite to causality, that is, there is no nrr-independence for $\hat{\varepsilon}_x = \mathrm{X}-\hat{f}_x(\mathrm{Y},\mathbf{S})$. However, they also showed that nrr-independence in both directions holds for a family of distributions $p(\mathrm{X},\mathrm{Y}|\mathbf{S})$ which, for $\mathbf{S}$ fixed, is characterized as the solutions of a third-order linear inhomogeneous differential equation. For example, Gaussian distributions belong to that family. Accordingly, if a system only contained AN equations with no hidden variables, an asymmetry $\hat{\varepsilon}_y \perp \mathrm{X}$ and $\hat{\varepsilon}_x \notperp \mathrm{Y}$, would suffice to infer a cause from $\mathrm{X}$ to $\mathrm{Y}$. This is because $\hat{\varepsilon}_y \perp \mathrm{X}$ always holds given the AN form of the functional equation of $\mathrm{Y}$ and only if the data generating distribution is within the special family of \cite{Hoyer2009} nrr-independence holds in both directions, in which case nothing can be concluded.

However, generally not all functional equations have an AN form. Focusing on the bivariate case, \cite{Janzing10} discussed the necessary assumptions for structure learning based on asymmetries of nrr-independencies. They indicated that, for a generative functional equation with the opposite direction of causality $\mathrm{X} = f_x(\mathrm{Y}, \varepsilon_x)$, it has to be assumed that $\hat{\varepsilon}_y \perp \mathrm{X}$ will not hold for any $\hat{\varepsilon}_y = \mathrm{Y}-\hat{f}_y(\mathrm{X})$, except within the family of \cite{Hoyer2009}. \cite{Janzing10} justified the fulfillment of this assumption because $\hat{\varepsilon}_y \perp \mathrm{X}$ would impose constraints making $p(\mathrm{Y})$ and $p(\mathrm{X}|\mathrm{Y})$ dependent. This dependence requires a fine tuning of the distribution $p(\mathrm{Y})$ of the cause, given the mechanism $p(\mathrm{X}|\mathrm{Y})$, and hence is fragile to changes in $p(\mathrm{Y})$ if the cause distribution changes independently of the causal mechanism, as expected. These arguments are tightly related with the justification of the faithfulness assumption for conditional independencies based on stability. In particular stability rules out 'pathological parameterizations' \citep{Pearl09} in which a conditional independence does not correspond to a d-separation present in the causal structure because such independencies also require tuning the parameters of the functional equation, and will vanish with small changes of these parameters.

When testing for nrr-independencies in multivariate systems, the common procedure starts by inferring a global causal ordering of the variables \citep{Mooij09}. This step already uses nrr-independencies, and relies on the fact that conditioning on a descendant introduces a dependence between $\mathrm{Y}$ and its noise variable. Subsequently, nrr-independencies are tested with regression models that, if the causal ordering is correct, do not take descendants as arguments. This allows removing superfluous edges from non-descendants that are not parents of $\mathrm{Y}$. To our knowledge, for the multivariate case an analogous assumption of faithfulness has not been formulated explicitly. For the sake of comparison with our results we here explicitly state the following assumption:

\vspace*{1mm}
\noindent \textbf{Assumption $\mathbf{1}$} \ \emph{Nrr-independence faithfulness for non-additive-noise functional equations}: `if the generative functional equation of $\mathrm{X}$, with $\mathrm{Y} \in \mathbf{\mathbf{Pa}}_x$, does not have an AN form, then $\hat{\varepsilon}_y \notperp \mathrm{X}, \forall \{\mathbf{S}, \hat{f}_y(\mathrm{X},\mathbf{S})\}$ with $\mathbf{S} \subseteq \mathbf{ND}(\mathrm{X})$, $\mathbf{\mathbf{Pa}}_x \backslash \mathrm{Y} \subseteq \mathbf{S}$ and $ \hat{\varepsilon}_y \equiv \mathrm{Y}-\hat{f}_y(\mathrm{X},\mathbf{S})$.'
%\vspace*{1mm}

This assumption is a multivariate version of the bivariate one discussed in \cite{Janzing10}. It considers that all other parents of $\mathrm{X}$ are included in the regression, and that only $\mathrm{X}$ and $\mathrm{Y}$ are exchanged. The assumption can be used iteratively when determining the causal ordering. It ensures that, if a functional equation does not have the AN form and hence Proposition $1$ does not guarantee independence in the right direction, an asymmetry of independence does not appear in the wrong direction. The assumption focuses on equations without an AN form because, by Proposition $1$, with the AN form nrr-independence in the wrong direction only leads to symmetric nrr-independencies.

Finally, we also review post-nonlinear AN models, where a global nonlinearity transforms the AN equation \citep{Zhang09}:
\begin{equation}
\label{e00}
\mathrm{V}_i = f_{i}(\mathbf{V}_i, \varepsilon_i) = h_{i,2}(h_{i,1}(\mathbf{V}_i)+ \varepsilon_i).
\end{equation}
Here $h_{i,2}$ is an invertible nonlinear function. For the bivariate case, with $\mathrm{Y} = h_2(h_1(\mathrm{X})+ \varepsilon_y)$, \cite{Zhang09} generalized the work of \cite{Hoyer2009} extending the characterization of the special family of distributions that admits a statistical post-nonlinear AN model in both directions. Furthermore, they showed how to fit a nonlinear model to extract residuals $\hat{\varepsilon}_y \equiv \hat{h}_2^{-1}(y) -\hat{h}_1(x)$ to test nrr-independencies. For the multivariate case, assuming no hidden variables, \cite{Zhang09} used regressions to evaluate nrr-independencies given sets of candidate parents previously determined examining conditional independencies between the variables \citep{Spirtes00}.

Additive-noise models are a well-established approach for structure learning, which has been mostly studied in the case of causal sufficiency (no hidden variables). A pure AN functional equation requires that the noise is separable as in Eq.\,\ref{e0}, and in the standard approach this property is exploited by testing the independence of the residuals of a nonlinear regression from the predictors. For multivariate systems, the application of these tests proceeds by inferring a global causal ordering. We extend the AN framework in four fronts, allowing for the presence of hidden variables, considering functional equations that have the AN form only after conditioning on certain variables, introducing an alternative regression-free test to infer causality, and modifying the procedure not to rely on the inference of a global causal ordering.

\section{Conditional variance independencies}
\label{s3}

%\begin{figure}[h]
%\vspace{1in}
%\caption{Sample Figure Caption}
%\end{figure}
We start introducing a regression-free test for causal directionality alternative to the regression-based analysis of nrr-independencies. For this purpose, we continue to consider the pure AN equations of the form of Eq.\,\ref{e0}. The key property of AN functional equations is that the independent noise $\varepsilon_i$ is additively separable from the parents. For a particular variable $\mathrm{Y}$ and a parent $\mathrm{X}$, define $\mathbf{Z} \equiv \mathbf{\mathbf{Pa}}_y \backslash \mathrm{X}$. We can study the conditional variance $\sigma_{\mathrm{Y}|\mathrm{X}, \mathbf{Z}}$ as a variable which is a function only of $\mathrm{X}$, with $\mathbf{Z}$ fixed. For AN functional equations, the independence and separability of the noise leads to $\sigma_{\mathrm{Y}|\mathrm{X}, \mathbf{Z}} = \sigma_{\varepsilon_y}$, which is independent of $\mathrm{X}$ ($\sigma_{\mathrm{Y}|\mathrm{X}, \mathbf{Z}} \perp \mathrm{X}$). This independence reflects the independence of the second-order moments of the residuals from the predictors indirectly, and does not require an actual reconstruction of the noise variables. Analogously to Proposition $1$, the AN form suffices for this type of independence, which we call conditional variance independence (cv-independence).

\vspace*{1mm}
\noindent \textbf{Proposition $\mathbf{2}$}\ \ \emph{Cv-independence with AN functional equations:} `If the functional equation of $\mathrm{Y}$ has the AN form, then $ \forall \mathrm{X} \in \mathbf{\mathbf{Pa}}_y \ \exists \mathbf{S}: \sigma_{\mathrm{Y}|\mathrm{X},\mathbf{S}} \perp \mathrm{X}$ $\forall \mathbf{S}= \mathbf{s}$.'

\vspace*{1mm}
The existence of at least one set $\mathbf{S}$ is guaranteed because $\mathbf{S} = \mathbf{Z}$ leads to $\sigma_{\mathrm{Y}|\mathrm{X},\mathbf{S}} \perp \mathrm{X}$. Because cv-independence follows from the fact that the noise is independent and separable from the other arguments of the equation, for the special family characterized by \cite{Hoyer2009} in which an AN statistical model can also be constructed in the reverse direction, cv-independence holds in both directions. For general systems, possibly containing functional equations without the AN form, an assumption analogous to Assumption $1$ is required to ensure that the asymmetry of cv-independencies does not hold in the direction inconsistent with the causal relation. Like for Assumption 1, to formulate this assumption of faithfulness we consider a functional equation with the opposite causal direction for $\mathrm{X}$ and $\mathrm{Y}$, and compare the cv-dependence of $\sigma_{\mathrm{Y}|\mathrm{X},\mathbf{S}}$ with respect to the independence stated in Proposition 2.

\vspace*{1mm}
\noindent \textbf{Assumption $\mathbf{2}$} \ \emph{Cv-independence faithfulness for non-additive-noise functional equations}: `if the functional equation of $\mathrm{X}$, with $\mathrm{Y} \in \mathbf{\mathbf{Pa}}_x$, does not have an AN form, then $\sigma_{\mathrm{Y}|\mathrm{X},\mathbf{S}} \notperp \mathrm{X}\ \forall \mathbf{S}\ \subseteq \mathbf{ND}(\mathrm{X}), \mathbf{\mathbf{Pa}}_x \backslash \mathrm{Y} \subseteq \mathbf{S}$.'
%\vspace*{1mm}

\vspace*{1mm}
The two faithfulness assumptions are related by the following conditions:

\vspace*{1mm}
\noindent \textbf{Proposition $3$} \ \emph{Relation between cv-independence faithfulness and nrr-independence faithfulness}: `The fulfillment of Assumption $2$ implies the one of Assumption $1$, but not the opposite.'

\vspace*{1mm}

\noindent Proof of Proposition $3$: See Appendix.

\vspace*{2mm}

Despite this theoretical asymmetry between the two faithfulness assumptions, the fulfilment of assumption $1$ and not assumption $2$ would impose further constraints to the probability distributions. It would require that $p(\hat{\varepsilon}_y|x,\mathbf{s})$ is such that dependencies appear in third or higher-order moments, so that cv-independence holds despite nrr-dependence. Furthermore, because we are considering a functional equation where $\mathrm{Y}$ is a parent of $\mathrm{X}$, the fulfillment of faithfulness regards $p(\hat{\varepsilon}_y|x,s)$, which does not correspond to the generative direction. Accordingly, cases in which nrr-independence faithfulness is violated and cv-independence faithfulness holds require a specific tuning introducing a dependence between the probability of the causes and the causal mechanism \citep{Janzing10}. The necessity of this tuning renders these cases fragile to changes in the distribution of the causes, and hence nonstable.

Testing nrr-independencies intrinsically requires a regression-based approach, fitting a (nonlinear) regression model. On the other hand, while cv-independencies can also be evaluated using the variance of the residuals, they can alternatively be tested in a regression-free approach, estimating the conditional variance of the variables without reconstructing the noise variables. The latter has the advantage that it does not rely on a particular model of regression. However, in some cases a test of variance homogeneity, if $\mathbf{S}$ is highly dimensional, may require more data than the nonlinear regression approach. These practical issues are out of the scope of this work. As we will see below, the cv-independencies formulation is particularly intuitive to derive an extension of AN models to partial CAN models.

\section{Conditionally-additive-noise models}
\label{s4}

For systems in which all functional equations have the AN form, full identifiability of the causal structure has been proven when there are no hidden variables \citep{Peters2014}. For partially AN models, for which only some of the equations have the AN form, asymmetries in nrr-independencies have been used \citep{Tillman09} as a method to complement algorithms of constraint-based causal discovery such as the PC algorithm \citep{Spirtes00}, which exploit conditional independencies between the variables. However, to our knowledge, it has not been examined how extra inferential power can be gained from functional equations that, although not having a pure AN form, are converted to the AN form after conditioning on some variables. We call these type of equations conditionally-additive-noise (CAN) functional equations. We derive the conditions on the form of a functional equation so that it can be converted to the CAN form in order to test nrr-independencies or cv-independencies. Furthermore, we now drop the assumption of causal sufficiency and consider also the existence of hidden variables.

To derive which equations have the conditionally-additive-noise form, we start expressing a generic functional equation as:
\begin{equation}
\label{e1}
\begin{split}
\mathrm{V}_i = f_i(\mathbf{V}_i,\varepsilon_i) = f_{i,1}(\mathrm{V}_{i,1,1},...,\mathrm{V}_{i,1,n_1}) + f_{i,2}(\mathrm{V}_{i,2,1},...,\mathrm{V}_{i,2,n_2},\varepsilon_i)+ f_{\varepsilon}(\varepsilon_i).
\end{split}
\end{equation}
Here the form allowed for $f_{i,1}$ and $f_{i,2}$ should be understood as complementary to simpler terms. That is, $f_{i,1}$ comprises any function of only $\mathbf{V}_i$. Function $f_{i,2}$ comprises any function that contains $\varepsilon_i$ as an argument, but excluding terms that only contain $\varepsilon_i$. The sets $\mathbf{V}_{i,1}$ and $\mathbf{V}_{i,2}$ can overlap. Any functional equation can be expressed in this form. In particular, if $\mathbf{V}_{i,2} = \emptyset$ the equation reduces to the AN form.

Consider $\mathrm{Y} = \mathrm{V}_i$ and a particular parent $\mathrm{X} \in \mathbf{V}_y$. We want to determine under which conditions cv-independencies or nrr-independencies can occur. As a first remark, if $\mathrm{X} \in \mathbf{V}_{y,2}$, $\sigma_{\mathrm{Y}|\mathrm{X},\mathbf{S}} \notperp \mathrm{X}$ for any set $\mathbf{S}$, since $\varepsilon_y$ is an argument of $f_{y,2}$ and $\mathrm{X}$ modulates its variance. For the same reason the residuals cannot be independent from $\mathrm{X}$ when $\mathrm{X} \in \mathbf{V}_{y,2}$. Subsequently, we focus on variables $\mathrm{X} \in \mathbf{V}_{y,1}$. Taking a particular variable $\mathrm{X}$ as reference, Eq.\,\ref{e1} can be expanded into the following subterms, where we also differentiate between observable variables (V) and hidden variables (U):
\begin{equation}
\label{e2}
\begin{split}
\mathrm{Y} &= f_{1,1}(\mathrm{X}, \mathbf{V}_{1,1}, \mathbf{U}_{1,1}) + f_{1,2}(\mathbf{V}_{1,2}, \mathbf{U}_{1,2}) + \sum_{j=1}^{n_{13}} f_{1,3,j}(\mathbf{\tilde{V}}_{1,3,j})\mathrm{V}_{1,3,j} + \sum_j \beta_j \mathrm{V}_{3,j}
\\&+\sum_{j=1}^{n_{14}} f_{1,4,j}(\mathbf{\tilde{V}}_{1,4,j}) \mathrm{U}_{1,4,j} + \sum_j \alpha_j \mathrm{U}_{3,j}
+ f_{2}(\mathbf{V}_{2}, \mathbf{U}_{2},\  \varepsilon_y) + f_{\varepsilon}(\varepsilon_y).
\end{split}
\end{equation}
We dropped subindex $y$ from all variables and functions to simplify the notation. As in Eq.\,\ref{e1}, the meaning of each function is determined by opposition to simpler terms explicitly separated. For example, $f_{1,1}$ is any function that does not have $\varepsilon_y$ as an argument and does not include the other explicit simpler terms that do not include $\varepsilon_y$ either. As will be appreciated below, we only separate those terms that are subject to different constraints in the conditions to obtain the CAN form. Only the function $f_{1,1}$ has $\mathrm{X}$ as an argument. Function $f_{1,3}$ is linear on some observable variables $\mathrm{V}_{1,3,j}$ with a coefficient that is a function $f_{1,3,j}$ of other observable variables. Function $f_{1,4}$ is linear in each hidden variable of $\mathbf{U}_{1,4}$, with a coefficient that is a function $f_{1,4,j}$ of observable variables $\mathbf{\tilde{V}}_{1,4}$. Here $\mathbf{\tilde{V}}_{1,3} = \{ \mathbf{\tilde{V}}_{1,3,1},...,\mathbf{\tilde{V}}_{1,3, n_{13}} \}$, and $\mathbf{\tilde{V}}_{1,4} = \{ \mathbf{\tilde{V}}_{1,4,1},...,\mathbf{\tilde{V}}_{1,4, n_{14}} \}$. Similarly $\mathbf{V}_{1,3} = \{ \mathrm{V}_{1,3,1},...,\mathrm{V}_{1,3,n_{13}} \}$ and $\mathbf{U}_{1,4} = \{ \mathrm{U}_{1,4,1},...,\mathrm{U}_{1,4,n_{14}} \}$. $\mathbf{V}_y = \{ \mathbf{V}_{1,1}, \mathbf{V}_{1,2}, \mathbf{V}_{1,3}, \mathbf{\tilde{V}}_{1,3} , \mathbf{\tilde{V}}_{1,4}, \mathbf{V}_{2}, \mathbf{V}_{3}\}$ contains all other observable parents apart from $\mathrm{X}$, and $\mathbf{U}_y = \{ \mathbf{U}_{1,1}, \mathbf{U}_{1,2}, \mathbf{U}_{1,4}, \mathbf{U}_{2}, \mathbf{U}_{3}\}$ all hidden parents. There can be overlaps between subgroups of $\mathbf{V}_y$ or of $\mathbf{U}_y$.

We determine the conditions that lead to cv-independencies and nrr-independencies. We will focus on the case in which, for a certain variable $\mathrm{X} \in \mathbf{PA}_y$, which causal relation with $\mathrm{Y}$ is examined, $\mathrm{X}$ is adjacent to all other potential causes of $\mathrm{Y}$, i.e.\,, parents and variables sharing a hidden common cause with $\mathrm{Y}$. This is because, as discussed above, it suffices that two observable potential causes are nonadjacent to infer that $\mathrm{Y}$ is a collider for them using conditional independencies \citep{Spirtes00}. This means that the conditions we derive could be relaxed, but the knowledge obtained would be redundant to the one provided by conditional independencies. Because cv-independencies only rely on second-order moments, there is a difference in the conditions needed to obtain cv-independence and nrr-independence. We start with cv-independencies, which lead to less restrictive conditions.

\subsection{The CAN form with cv-independence}
%\vspace*{2mm}

We define the cv-CAN form as the form of a functional equation leading to cv-independence:

\vspace*{1mm}
\noindent \textbf{Definition $1$}\ \ \emph{Cv-independence with cv-CAN functional equations:} `The functional equation of $\mathrm{Y}$ has the cv-CAN form for $\mathrm{X}$ when conditioning on $\mathbf{S}$ if $\sigma_{\mathrm{Y}|\mathrm{X},\mathbf{S}} \perp \mathrm{X}$ $\forall \mathbf{S}= \mathbf{s}$.'

We now enunciate when a functional equation can be set into the cv-CAN form. For this purpose, expressing the functional equation of $\mathrm{Y}$ as in Eq.\,\ref{e2}, we define the functions $\mathrm{Y}_{1,2} \equiv f_{1,2}(\mathbf{U}_{1,2};\mathbf{V}_{1,2})$ and $\mathrm{Y}_2 \equiv f_{2}(\mathbf{U}_{2}, \varepsilon_y;\mathbf{V}_{2})$, where $\mathbf{V}_{1,2}$ and $\mathbf{V}_{2}$ play the role of fixed parameters, and we also define $\mathbf{S}_2 \equiv \{ \mathrm{Y}_{1,2}, \mathrm{Y}_2 , \mathbf{U}_{1,4} , \mathbf{U}_{3} \}$. The cv-CAN form is characterized as follows.

\vspace*{1mm}
\noindent \textbf{Theorem $\mathbf{1}$}\ \ \emph{Functional equations with the cv-CAN form:} `Consider an $\mathrm{X} \in \mathbf{Pa}_y$ and a set $\mathbf{S}$. For the case in which $\mathrm{X}$ is adjacent to all other potential causes of $\mathrm{Y}$, the functional equation of $\mathrm{Y}$ has the cv-CAN form with respect to $\mathrm{X}$ given the set $\mathbf{S}$ if and only if the hidden variables fulfill the following conditions
%\begin{equation}
%\label{e2b}
%\begin{split}
%&\mathrm{i)}\ \mathbf{U}_{1,1} = \emptyset; \ \ \ \ \ \ \ \ \ \ \ \ \ \ \ \ \ \ \ \ \ \ \ \ \ \ \ \ \ \ \ \ \ \ \ \ \ \ \ \ \ \ \ \ \ \ \ \ \ \mathrm{ii)}\ \mathrm{X} \perp \mathrm{U}_k|\mathbf{S}\ \forall \mathrm{U}_k \in \{\mathbf{U}_{1,2} , \mathbf{U}_{2}\}; \\ &\mathrm{iii)}\ \sigma_{\mathrm{U}_k|\mathrm{X},\mathbf{S}} \perp \mathrm{X} \ \forall \mathrm{U}_k \in \{\mathbf{U}_{1,4} , \mathbf{U}_{3}\}; \ \ \ \ \ \ \ \ \ \ \ \ \mathrm{iv)}\ \sigma_{\mathrm{Z}_i\mathrm{Z}_j|\mathrm{X},\mathbf{S}} \perp \mathrm{X}\ \forall \mathrm{Z}_i, \mathrm{Z}_j \in \mathbf{S}_2,
%\end{split}
%\end{equation}
\begin{equation}
\label{e2b}
\begin{split}
&\mathrm{i)}\ \mathbf{U}_{1,1} = \emptyset; \\ &\mathrm{ii)}\ \mathrm{X} \perp \mathrm{U}_k|\mathbf{S}\ \forall \mathrm{U}_k \in \{\mathbf{U}_{1,2} , \mathbf{U}_{2}\}; \\ &\mathrm{iii)}\ \sigma_{\mathrm{U}_k|\mathrm{X},\mathbf{S}} \perp \mathrm{X} \ \forall \mathrm{U}_k \in \{\mathbf{U}_{1,4} , \mathbf{U}_{3}\}; \\ &\mathrm{iv)}\ \sigma_{\mathrm{Z}_i\mathrm{Z}_j|\mathrm{X},\mathbf{S}} \perp \mathrm{X}\ \forall \mathrm{Z}_i, \mathrm{Z}_j \in \mathbf{S}_2,
\end{split}
\end{equation}
the set $\mathbf{S}$ is such that
\begin{equation}
\label{e2c}
\begin{split}
\{ \mathbf{V}_{1,1} , \mathbf{V}_{1,2}, \mathbf{\tilde{V}}_{1,3}, \mathbf{\tilde{V}}_{1,4}, \mathbf{V}_{2}, \mathbf{V}_{1,3,2}, \mathbf{V}_{3,2} \} \subseteq \mathbf{S} \notag,
\end{split}
\end{equation}
where $\mathbf{V}_{3,2}$ is defined as $\mathbf{V}_{3,2} \equiv \mathbf{V}_{3} \backslash \mathbf{V}_{3,1}$, with $\mathbf{V}_{3,1} \subseteq \mathbf{V}_{3}$ such that $\forall \mathrm{V}_i \in \mathbf{V}_{3,1}\ \sigma_{\mathrm{V}_i|\mathrm{X}, \mathbf{S}} \perp \mathrm{X} $, $\mathbf{V}_{1,3,2}$ is defined as $\mathbf{V}_{1,3,2} \equiv \mathbf{V}_{1,3} \backslash \mathbf{V}_{1,3,1}$ with $\mathbf{V}_{1,3,1} \subseteq \mathbf{V}_{1,3}$ such that $\forall \mathrm{V}_i \in \mathbf{V}_{1,3,1}\ \sigma_{\mathrm{V}_i|\mathrm{X}, \mathbf{S}} \perp \mathrm{X}$, and the unconditioned observable variables also fulfill the following conditions
%\begin{equation}
%\label{e2d}
%\begin{split}
%\mathrm{v)}\ \sigma_{\mathrm{V}_i\mathrm{V}_j|\mathrm{X},\mathbf{S}} \perp \mathrm{X}\ \forall \mathrm{V}_i, \mathrm{V}_j \in \mathbf{S}_3;\ \ \ \ \ \ \ \ \ \ \ \
%\mathrm{vi)}\ \sigma_{\mathrm{V}_i\mathrm{Z}_j|\mathrm{X},\mathbf{S}} \perp \mathrm{X}\ \forall \mathrm{V}_i \in \mathbf{S}_3, \mathrm{Z}_j \in \mathbf{S}_2,
%\end{split}
%\end{equation}
\begin{equation}
\label{e2d}
\begin{split}
&\mathrm{v)}\ \sigma_{\mathrm{V}_i\mathrm{V}_j|\mathrm{X},\mathbf{S}} \perp \mathrm{X}\ \forall \mathrm{V}_i, \mathrm{V}_j \in \mathbf{S}_3;\\ &\mathrm{vi)}\ \sigma_{\mathrm{V}_i\mathrm{Z}_j|\mathrm{X},\mathbf{S}} \perp \mathrm{X}\ \forall \mathrm{V}_i \in \mathbf{S}_3, \mathrm{Z}_j \in \mathbf{S}_2,
\end{split}
\end{equation}
where $\mathbf{S}_3 = \{ \mathbf{V}_{3,1}, \mathbf{V}_{1,3,1}\}$.'
\vspace*{1mm}

\noindent Proof of Theorem $1$: See Appendix.

\vspace*{2mm}

To understand the logic of these conditions, we rewrite Eq.\,\ref{e2} as
\begin{equation}
\label{e3}
\begin{split}
%\mathrm{Y} &= f_{1,1}(\mathrm{X} ;\mathbf{V}_{1,1})+ \left [
%f_{1,2}(\mathbf{U}_{1,2};\mathbf{V}_{1,2}) \right.\\&+ \sum_j \tilde{\beta}_j \mathrm{V}_{1,3,1,j} + \sum_j \beta_j \mathrm{V}_{3,1,j}
%\\&+\sum_j \tilde{\alpha}_j \mathrm{U}_{1,4,j} + \sum_j \alpha_j \mathrm{U}_{3,j}
%\\&+ \left. f_{2}(\mathbf{U}_{2}, \varepsilon_y;\mathbf{V}_{2}) +  f_{\varepsilon}(\varepsilon_y) + c \right ].
\mathrm{Y} &= f_{1,1}(\mathrm{X} ;\mathbf{V}_{1,1})+ \left [
f_{1,2}(\mathbf{U}_{1,2};\mathbf{V}_{1,2}) + \sum_j \tilde{\beta}_j \mathrm{V}_{1,3,1,j} + \sum_j \beta_j \mathrm{V}_{3,1,j} \right.\\&+\left. \sum_j \tilde{\alpha}_j \mathrm{U}_{1,4,j} + \sum_j \alpha_j \mathrm{U}_{3,j} +  f_{2}(\mathbf{U}_{2}, \varepsilon_y;\mathbf{V}_{2}) +  f_{\varepsilon}(\varepsilon_y) + c \right ].
\end{split}
\end{equation}
$\tilde{\beta}_j = f_{1,3,1,j}(\mathbf{\tilde{V}}_{1,3,1,j})$ and $\tilde{\alpha}_j = f_{1,4,j}(\mathbf{\tilde{V}}_{1,4,j})$ are constant coefficients because $\{ \mathbf{\tilde{V}}_{1,3}, \mathbf{\tilde{V}}_{1,4}\} \subseteq \mathbf{S} $. The constant $c$ equals
$\sum_j \tilde{\beta}_j \mathrm{V}_{1,3,2,j} + \sum_j \beta_j \mathrm{V}_{3,2,j}$ because $\{ \mathbf{\tilde{V}}_{1,3}, \mathbf{V}_{1,3,2}, \mathbf{V}_{3,2} \} \subseteq \mathbf{S} $. Eq.\,\ref{e3} can be summarized as:
\begin{equation}
\label{e4}
\begin{split}
\mathrm{Y} = f_{1,1}(\mathrm{X} ;\mathbf{S}) + g(\mathbf{V}_{1,3,1}, \mathbf{V}_{3,1},\mathbf{U}_{1,2}, \mathbf{U}_{1,4}, \mathbf{U}_{2}, \mathbf{U}_{3};\mathbf{S}) = f_{1,1}(\mathrm{X} ;\mathbf{S}) + \xi_{y|\mathbf{S}},
\end{split}
\end{equation}
where the function $g$ plays the role of a noise $\xi_{y|\mathbf{S}}$ analogous to the additive noise term of a pure AN equation, and hence the equation has the additive-noise form when seen as a function of $\mathrm{X}$. $\mathbf{U}_{1,2}$ and $\mathbf{U}_{2}$ are conditionally independent of $\mathrm{X}$ given $\mathbf{S}$, and $g$ is linear in all the other arguments, with their variances and covariances conditionally independent of $\mathrm{X}$ given $\mathbf{S}$. This leads to $\sigma_{\xi_{y|\mathbf{S}}|\mathrm{X},\mathbf{S}} \perp \mathrm{X}$. Note that, to fulfill the conditions in Eqs.\,\ref{e2b} and \ref{e2d}, $\mathbf{S}$ may need to include other variables that are not parents of $\mathrm{Y}$. Furthermore, the constraints are intertwined because independencies change depending on which variables are included in $\mathbf{S}$. Since all variables in $\mathbf{V}_{3}$ and $\mathbf{V}_{1,3}$ are observable, it is always possible to try to find a valid set $\mathbf{S}$ with $\mathbf{V}_{3,1} = \emptyset$ and $\mathbf{V}_{1,3,1} = \emptyset$. In that case, the constraints of Eq.\,\ref{e2d} vanish. It is also possible to formulate a simpler sufficient condition by demanding $\mathrm{X} \perp \mathrm{U}_k |\mathbf{S} \ \forall \mathrm{U}_k \in \{ \mathbf{U}_y \backslash \mathbf{U}_{1,1}\}$.

Note that the cv-CAN form is obtained relative to a certain variable. The existence of a valid set $\mathbf{S}$ to place an equation in the CAN form relative to a variable is not guaranteed for all the observable parents. This is because of two reasons. First, it may be due to the presence of hidden variables that for a certain $\mathrm{X}$ do not fulfill the conditions of Theorem $1$. This limitation is common to pure AN functional equations if hidden variables are allowed, since AN equations are CAN equations with $\mathbf{V}_{y,2} = \emptyset$. Second, even with no hidden variables, $\forall \mathrm{V}\ \in \mathbf{V}_{y,2}\ \nexists \mathbf{S}: \sigma_{\mathrm{Y}|\mathrm{V},\mathbf{S}} \perp \mathrm{V}$. That is, certain parents are not additively separable from the noise and cannot lead to any cv-independence. The fact that only some equations in the system, and only relatively to certain variables, have the CAN form, hinders the application of algorithms of structure learning in which a global causal ordering is inferred searching for the ordering that leads to the highest estimates of residuals independence \citep{Mooij09, Peters2014}, which are designed for systems in which all equations have the pure AN form. This is because now a lack of independence can be due not to the wrong order, but to the lack of separability of the noise, for the reasons mentioned above.

Theorem 1 states which form of a functional equation will create a cv-independence. Assuming that a certain functional equation is known or hypothesized, and for a certain context in which the existence of certain hidden variables is known or hypothesized, the theorem allows determining if a cv-independence exists. However, the theorem cannot be applied for inference, given that the conditions in Eqs.\,\ref{e2b} and \ref{e2d} involve hidden variables and hence their fulfillment cannot be tested from data. To derive a criterion applicable for inference, we identify the assumptions required so that a specific asymmetry of cv-independencies provides information about the causal relation between the corresponding pair of variables, without inferring a global causal ordering.

\vspace*{1mm}
\noindent \textbf{Assumption $3$} \ \emph{Cv-independence faithfulness for non-conditionally-additive-noise functional equations}: `if $\forall \mathbf{S}'\ \subseteq \mathbf{S}$ the generative functional equation of $\mathrm{X}$, with $\mathrm{Y} \in \mathbf{\mathbf{Pa}}_x$, does not have the cv-CAN form for $\mathrm{Y}$ conditioned on $\mathbf{S}'$, then $\ \sigma_{\mathrm{Y}|\mathrm{X},\mathbf{S}} \notperp \mathrm{X}$.'
\vspace*{1mm}

In comparison to the previous assumptions of faithfulness, here there is no restriction of $\mathbf{S}$ to non-descendants of $\mathrm{X}$. $\mathbf{S}$ is not limited based on any causal knowledge. The assumption again focuses on functional equations which do not have the CAN form. In the Appendix we indicate that, like for pure AN equations, a special family of joint distributions $p(\mathrm{X},\mathrm{Y}|\mathbf{S})$ as described by \cite{Hoyer2009}, allows a CAN statistical form in both directions. Assumption $3$ can be used to infer a potential cause from $\mathrm{X}$ to $\mathrm{Y}$, that is, to infer that $\mathrm{X}$ causes $\mathrm{Y}$ or there is a latent common cause:

\vspace*{1mm}
\noindent \textbf{Proposition $4$}\ \ \emph{Inferring noncausality with cv-independence asymmetries}: `Consider two adjacent variables $\mathrm{X}$ and $\mathrm{Y}$. Under the assumption of cv-independence faithfulness for non-cv-CAN functional equations (Assumption $3$), if $\exists \mathbf{S}: \sigma_{\mathrm{Y}|\mathrm{X}, \mathbf{S}} \perp \mathrm{X}$ and $\sigma_{\mathrm{X}|\mathrm{Y}, \mathbf{S}'} \notperp \mathrm{Y}\ \forall \mathbf{S}'\ \subseteq \mathbf{S}$, then there is no causality from $\mathrm{Y}$ to $\mathrm{X}$, that is, $\mathrm{X}$ is a potential cause of $\mathrm{Y}$.'

\vspace*{1mm}

\noindent Proof of Proposition $4$: If $\exists \mathbf{S}: \sigma_{\mathrm{Y}|\mathrm{X}, \mathbf{S}} \perp \mathrm{X}$, it does not hold that $\sigma_{\mathrm{Y}|\mathrm{X},\mathbf{S}} \notperp \mathrm{X}$. By Assumption $3$, this implies that, either $\mathrm{Y} \notin \mathbf{\mathbf{Pa}}_x$ or the functional equation of $\mathrm{X}$ has the cv-CAN form for $\mathrm{Y}$ conditioning on $\mathbf{S}'$ for some $\mathbf{S}'\ \subseteq \mathbf{S}$. The latter is discarded since we have $\sigma_{\mathrm{X}|\mathrm{Y}, \mathbf{S}'} \notperp \mathrm{Y}\ \forall \mathbf{S}'\ \subseteq \mathbf{S}$. $\ \ \Box$
%\noindent Proof of Proposition $5$: See Appendix.

\vspace*{2mm}
We now provide some intuition about this criterion. First, if there is only a latent common cause between $\mathrm{X}$ and $\mathrm{Y}$, it is valid to infer a potential cause in either direction. Therefore, what we need is to avoid inferring the potential cause in the wrong direction when there is a genuine cause. For the bivariate case, the asymmetry of cv-independencies suffices if we assume faithfulness for non-CAN functional equations. However, conditioning on some set $\mathbf{S}$ not only can convert an equation to the CAN form, it can also introduce cv-dependencies that were not present when conditioning only on a subset of $\mathbf{S}$. An asymmetry could appear in the following way: for a certain $\mathbf{S}^*$, not only the functional equation of $\mathrm{Y}$ has the CAN form relatively to $\mathrm{X}$, but furthermore the conditional joint distribution $p(\mathrm{X},\mathrm{Y}|\mathbf{S}^*)$ belongs to the special family that allows a CAN statistical model in both directions. For $\mathbf{S}^*$, a symmetry of cv-independencies is obtained. However, conditioning on a larger set ($\mathbf{S}^* \subset \mathbf{S}$) can introduce a cv-dependence that only appears in the direction in which the independence given $\mathbf{S}^*$ was consistent with the causal structure. Accordingly, for $\mathbf{S}$ an unfaithful asymmetry is obtained. See Section \ref{s5} for an example of a system in which this type unfaithful of asymmetry occurs. Checking if $\sigma_{\mathrm{X}|\mathrm{Y}, \mathbf{S}'} \notperp \mathrm{Y}\ \forall \mathbf{S}'\ \subseteq \mathbf{S}$, we can find the $\mathbf{S}^* \subset \mathbf{S}$ for which symmetric independencies were obtained, showing that the observed asymmetry is not reliable.

Altogether, Theorem 1 states when cv-independencies occur as a consequence of the causal structure, and Assumption 3 specifies the faithfulness assumption required so that cv-independencies do not occur inconsistently with the causal structure, which allows formulating the criterion of Proposition $4$ to infer noncausality from data. That is, Theorem 1 provides us an analytical tool to establish cv-independencies from a known or hypothesized functional equation, and Proposition $4$ provides us an empirical tool to infer the causal information from data.

\subsection{The CAN form with nrr-independence}

We now define the nrr-CAN form as the form of a functional equation leading to nrr-independence:

\vspace*{1mm}
\noindent \textbf{Definition $2$}\ \ \emph{Nrr-independence with nrr-CAN functional equations:} `The functional equation of $\mathrm{Y}$ has the nrr-CAN form for $\mathrm{X}$ when conditioning on $\mathbf{S}$ if $\forall \mathbf{S}= \mathbf{s}$ $\exists \hat{f}_y(\mathrm{X};\mathbf{S})$ such that $\hat{\varepsilon}_{y|\mathrm{X};\mathbf{S}} \perp \mathrm{X}$, with $\hat{\varepsilon}_{y|\mathrm{X};\mathbf{S}} \equiv \mathrm{Y}-\hat{f}_y(\mathrm{X};\mathbf{S})$.'

We distinguish $\mathrm{X}$ and $\mathbf{S}$ as an argument and constant parameters of the function $\hat{f}_y$, since $\mathbf{S}= \mathbf{s}$ is fixed when conditioning. We now enunciate the conditions in which a functional equation can be set into the nrr-CAN form. Similarly to Theorem $1$, we focus on conditions for the case that $\mathrm{X}$ is adjacent to all other potential causes of $\mathrm{Y}$, since otherwise the rules based on conditional independencies would already be applicable to extract the same causal information. For this purpose, we first introduce some further notation. Consider a variable $\mathrm{Z} \in \{\mathbf{V}_{3}, \mathbf{U}_{3} \}$. This variable has a linear additive contribution to the functional equation of $\mathrm{Y}$ (Eq.\,\ref{e2}), and hence $\hat{f}_y(\mathrm{X};\mathbf{S})$ will contain an additive component associated with the term in which $\mathrm{Z}$ appears. This component corresponds to the conditional mean of $\mathrm{Z}$ given $\mathrm{X}$ and $\mathbf{S}$, scaled by its coefficient in Eq.\,\ref{e2}. The contribution of this term to the residual of $\mathrm{Y}$ is hence proportional to the residual $\varepsilon_{z|\mathrm{X};\mathbf{S}}$ that would result from a separate regression to estimate $\mathrm{Z}$. Therefore, we define $\varepsilon_{z|\mathrm{X};\mathbf{S}} \equiv \mathrm{Z}- \hat{f}_{z}(\mathrm{X};\mathbf{S})$ for $\mathrm{Z} \in \{\mathbf{V}_{3}, \mathbf{U}_{3} \}$. We use an analogous definition in relation to the part of the residual of $\mathrm{Y}$ associated with $\mathrm{Z} \in \{\mathbf{V}_{1,3} , \mathbf{U}_{1,4}\}$ when, after conditioning on $\mathbf{\tilde{V}}_{1,3}$ and $\mathbf{\tilde{V}}_{1,4}$, respectively, they also have linearly additive contributions in Eq.\,\ref{e2}. The nrr-CAN form is characterized as follows:

\vspace*{1mm}
\noindent \textbf{Theorem $\mathbf{2}$}\ \ \emph{Functional equations with nrr-CAN form:} `Consider an $\mathrm{X} \in \mathbf{Pa}_y$ and the case in which $\mathrm{X}$ is adjacent to all other potential causes of $\mathrm{Y}$. Express the functional equation of $\mathrm{Y}$ as in Eq.\,\ref{e2}. The equation has the nrr-CAN form with respect to $\mathrm{X}$ given $\mathbf{S}$ if and only if the hidden variables fulfill the following conditions
%\begin{equation}
%\label{e2b}
%\begin{split}
%&\mathrm{i)}\ \mathbf{U}_{1,1} = \emptyset; \ \ \ \ \ \ \ \ \ \ \ \ \ \ \ \ \ \ \ \ \ \ \ \ \ \ \ \ \ \ \ \ \ \ \ \ \ \ \ \ \ \ \ \ \ \ \ \ \ \mathrm{ii)}\ \mathrm{X} \perp \mathrm{U}_k|\mathbf{S}\ \forall \mathrm{U}_k \in \{\mathbf{U}_{1,2} , \mathbf{U}_{2}\}; \\ &\mathrm{iii)}\ \varepsilon_{\mathrm{U}_k|\mathrm{X},\mathbf{S}} \perp \mathrm{X} \ \forall \mathrm{U}_k \in \{\mathbf{U}_{1,4} , \mathbf{U}_{3}\},
%\end{split}
%\end{equation}
\begin{equation}
\label{e2b2}
\begin{split}
&\mathrm{i)}\ \mathbf{U}_{1,1} = \emptyset; \\ &\mathrm{ii)}\ \mathrm{X} \perp \mathrm{U}_k|\mathbf{S}\ \forall \mathrm{U}_k \in \{\mathbf{U}_{1,2} , \mathbf{U}_{2}\}; \\ &\mathrm{iii)}\ \varepsilon_{\mathrm{U}_k|\mathrm{X}; \mathbf{S}} \perp \mathrm{X} \ \forall \mathrm{U}_k \in \{\mathbf{U}_{1,4} , \mathbf{U}_{3}\},
\end{split}
\end{equation}

the set $\mathbf{S}$ is such that
\begin{equation}
\label{e2c2}
\begin{split}
\{ \mathbf{V}_{1,1} , \mathbf{V}_{1,2}, \mathbf{\tilde{V}}_{1,3}, \mathbf{\tilde{V}}_{1,4}, \mathbf{V}_{2}, \mathbf{V}_{1,3,2}, \mathbf{V}_{3,2} \} \subseteq \mathbf{S} \notag,
\end{split}
\end{equation}
where $\mathbf{V}_{3,2}$ is defined as $\mathbf{V}_{3,2} \equiv \mathbf{V}_{3} \backslash \mathbf{V}_{3,1}$, with $\mathbf{V}_{3,1} \subseteq \mathbf{V}_{3}$ such that $\forall \mathrm{V}_i \in \mathbf{V}_{3,1}\ \varepsilon_{\mathrm{V}_i|\mathrm{X}; \mathbf{S}} \perp \mathrm{X} $, $\mathbf{V}_{1,3,2}$ is defined as $\mathbf{V}_{1,3,2} \equiv \mathbf{V}_{1,3} \backslash \mathbf{V}_{1,3,1}$ with $\mathbf{V}_{1,3,1} \subseteq \mathbf{V}_{1,3}$ such that $\forall \mathrm{V}_i \in \mathbf{V}_{1,3,1}\ \varepsilon_{\mathrm{V}_i|\mathrm{X}; \mathbf{S}} \perp \mathrm{X}$.

%\begin{equation}
%\label{e2f}
%\begin{split}
%&\mathbf{U}_{1,1} = \emptyset,\\
%&\mathrm{X} \perp \mathrm{U}_k|\mathbf{S}\ \forall \mathrm{U}_k \in \mathbf{U}_y \backslash \mathbf{U}_{1,1}.
%\end{split}
%\end{equation}
%
%\vspace*{1mm}

\noindent Proof of Theorem $2$: See Appendix.

\vspace*{1mm}
The correspondence between Theorems $1$ and $2$ can be understood considering that the conditional variances only quantify, in a regression-free way, dependencies of $\mathrm{X}$ with the second-order moments of the residuals of $\mathrm{Y}$. On the other hand, nrr-independencies are sensitive also to dependencies of $\mathrm{X}$ with the residuals higher-order moments. Accordingly, while the conditions i-ii) of Theorem $1$ requiring conditional independencies are preserved in Theorem $2$, the rest of conditions iii-vi), specific for second-order moments, are modified. Condition iii) of Theorem $2$ is analogous to condition iii) of Theorem $1$. It indicates that for $\mathrm{U} \in \{\mathbf{U}_{1,4} , \mathbf{U}_{3}\}$ a dependence with $\mathrm{X}$ can exist in the mean $\mu_{u|\mathrm{X};\mathbf{S}}$, which will be captured by the regression function, but any other dependence with $\mathrm{X}$ in $\varepsilon_{u|\mathrm{X};\mathbf{S}}$ will create also an nrr-dependence between $\mathrm{X}$ and the residuals $\varepsilon_{y|\mathrm{X};\mathbf{S}}$. The other conditions of Theorem $1$, iv-vi), are already fulfilled given the standard assumption of faithfulness for conditional independencies \citep{Spirtes00}. This because in Theorem $1$ condition iii) and the requirements in the selection of $\mathbf{V}_{3,1}$ and $\mathbf{V}_{1,3,1}$ only involve conditional variances, and the conditional variance of $\mathrm{Y}$ also depends on the covariance between the different linear contributions in its functional equation. Conversely, in Theorem $2$ condition iii) and the requirements in the selection of $\mathbf{V}_{3,1}$ and $\mathbf{V}_{1,3,1}$ are conditional independence constraints. Any dependence between $\mathrm{X}$ and a subset of variables in $\mathbf{S}_2$ or $\mathbf{S}_3$ which exists despite $\mathrm{X}$ being independent of each of these single variables would violate the standard assumption of faithfulness for conditional independencies.

For most functional equations, both or none of the CAN forms are obtainable, because the existence of higher-order dependencies without second-order dependencies imposes restrictive constraints to the form of the functional equations.
However, the specific cases in which the cv-CAN form holds and the nrr-CAN form does not may still be stable, in the sense that they do not depend on a specific tuning of the distribution of the causes \citep{Janzing10}. This is because the independencies required in Theorem 1 and 2 may depend exclusively on the form of the functional equations. The relation between the fulfillment of the cv-CAN form and the nrr-CAN form is thus qualitatively different than the one of cv-independence faithfulness and nrr-independence faithfulness, as discussed in relation to Proposition 3. In the latter case, because the violation of faithfulness regards dependencies with residuals extracted in the direction opposite to the generative functional equation, cases in which cv-independence faithfulness is violated and nrr-independence faithfulness is not will occur only for specific tunings of the distribution of the causes, as discussed above.

Similarly to the formulation based on cv-independencies, the conditions in Eq.\,\ref{e2b} are not testable experimentally, since they involve hidden variables. Again, Theorem 2 serves to identify for which type of functional equations nrr-independencies will exist as a consequence of the form of the equation, but furthermore a criterion for inference from data has to be introduced. For this purpose we formulate for nrr-independence an assumption of faithfulness analogous to Assumption $3$:

\vspace*{1mm}
\noindent \textbf{Assumption $4$} \ \emph{Nrr-independence faithfulness for non-conditionally-additive-noise functional equations}: `if $\forall \mathbf{S}'\ \subseteq \mathbf{S}$ the generative functional equation of $\mathrm{X}$, with $\mathrm{Y} \in \mathbf{\mathbf{Pa}}_x$, does not have the nrr-CAN form for $\mathrm{Y}$ conditioned on $\mathbf{S}'$, then $\hat{\varepsilon}_{y|\mathrm{X};\mathbf{S}} \notperp \mathrm{X}$\ for any regression $\hat{f}_y(\mathrm{X};\mathbf{S})$, with $\hat{\varepsilon}_{y|\mathrm{X};\mathbf{S}} \equiv \mathrm{Y}-\hat{f}_y(\mathrm{X};\mathbf{S})$.'
\vspace*{1mm}

Based on this assumption, we can state a criterion of noncausality using nrr-independencies analogous to Proposition $4$:

\vspace*{1mm}
\noindent \textbf{Proposition $5$}\ \ \emph{Inferring noncausality with nrr-independence asymmetries}: `Under the assumption of nrr-independence faithfulness for non-nrr-CAN functional equations (Assumption $4$), if $\exists \mathbf{S}$ and $\hat{f}_y(\mathrm{X};\mathbf{S}): \hat{\varepsilon}_{y|\mathrm{X};\mathbf{S}} \perp \mathrm{X}$ with $\hat{\varepsilon}_{y|\mathrm{X};\mathbf{S}} \equiv \mathrm{Y}-\hat{f}_y(\mathrm{X};\mathbf{S})$
 and $\hat{\varepsilon}_{x|\mathrm{Y};\mathbf{S}'} \notperp \mathrm{Y}$\ for any regression $\hat{f}_x(\mathrm{Y};\mathbf{S}'), \forall \mathbf{S}'\ \subseteq \mathbf{S}$, with $\hat{\varepsilon}_{x|\mathrm{Y};\mathbf{S}'} \equiv \mathrm{X}-\hat{f}_x(\mathrm{Y};\mathbf{S}')$, then there is no causality from $\mathrm{Y}$ to $\mathrm{X}$.'

\vspace*{1mm}

\noindent Proof of Proposition $5$: If $\exists \mathbf{S}$ and $\hat{f}_y(\mathrm{X};\mathbf{S}): \hat{\varepsilon}_{y|\mathrm{X};\mathbf{S}} \perp \mathrm{X}$ with $\hat{\varepsilon}_{y|\mathrm{X};\mathbf{S}} \equiv \mathrm{Y}-\hat{f}_y(\mathrm{X};\mathbf{S})$, it does not hold that $\hat{\varepsilon}_{y|\mathrm{X};\mathbf{S}} \notperp \mathrm{X}$ for any $\hat{f}_y(\mathrm{X};\mathbf{S})$. By assumption $4$, this implies that, either $\mathrm{Y} \notin \mathbf{\mathbf{Pa}}_x$ or the functional equation of $\mathrm{X}$ has the nrr-CAN form for $\mathrm{Y}$ conditioning on $\mathbf{S}'$ for some $\mathbf{S}'\ \subseteq \mathbf{S}$. The latter is discarded since we have $\hat{\varepsilon}_{x|\mathrm{Y};\mathbf{S}'} \notperp \mathrm{Y}$\ for any regression $\hat{f}_x(\mathrm{Y};\mathbf{S}'), \forall \mathbf{S}'\ \subseteq \mathbf{S}$.$\ \ \Box$

\vspace*{2mm}
This criterion is analogous to the one with cv-independencies. However, because nrr-independencies are a regression-based approach, there is an extra condition requiring that dependencies hold for any possible regression. Theoretically, this is an extra requirement to apply nrr-independencies for causal discovery as opposed to cv-independencies. Pragmatically, this reduces to the requirement of a good regression model, in the same way that we need a good estimate of the conditional variances. Note that the use of nonlinear regressions differs from that common in algorithms that infer a global causal ordering \citep{Mooij09}. In that approach, a regression takes as predictors all the candidate parents of a variable. Conversely, here the regression operates on $\mathrm{X}$ with all variables in $\mathbf{S}$ conditioned, or at least, regarding the terms in Eq.\,\ref{e3}, it has to estimate $f_{1,1}$ as a function of $\mathrm{X}$ and the subset of variables in $\mathbf{V}_{1,1}$ which does not appear in any other term, while conditioning on the rest. The relation between a formulation of nrr-independence in terms of conditional regressions and of multivariate regressions will be further addressed in future work.

\section{Examples}
\label{s5}

%We first examine concrete systems with certain causal structures and functional equations. We later examine, as a proof of concept, the applicability of the criteria of causal inference to infer the causal structure of a set of systems from which we generate simulated data.

%\subsection{Concrete systems}

\begin{figure}[t]
  \vspace{0.15in}
  \begin{center}
    \scalebox{0.6}{\includegraphics*{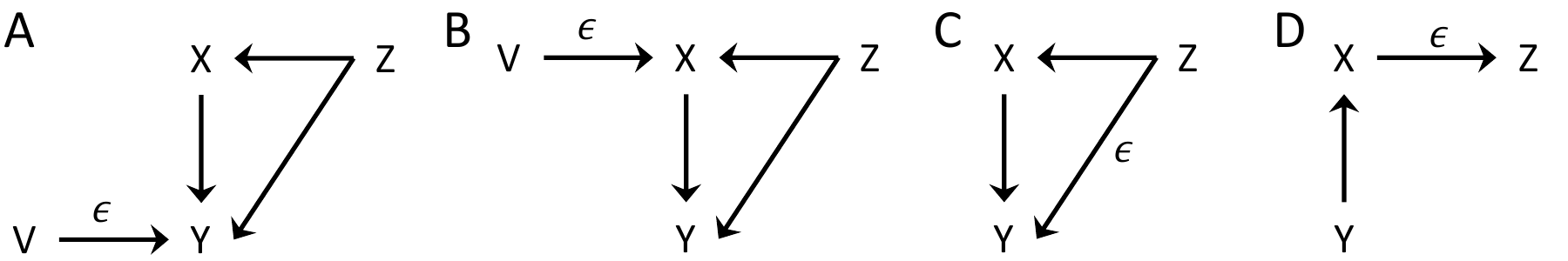}}
  \end{center}
  \vspace{0.25in}
  \caption{Examples of the effect of conditioning on cv-independencies. The examples represent Gaussian Linear Mixed Models (Eq.\,\ref{elm}), with random coefficients indicated by $\xrightarrow[]{\epsilon}$. The corresponding cv-dependencies between $\mathrm{X}$ and $\mathrm{Y}$ conditioned or unconditioned on $\mathrm{Z}$ are collected in Table \ref{t1}. Systems with the same causal structure of A) and B) and more general functional equations are described in Eqs.\,\ref{ee1}-\ref{ee3}.}
  \label{f1}
\end{figure}

We now examine some concrete examples to understand the different possible effects that conditioning on an extra variable can have to confer the CAN form or remove it from a functional equation. For that purpose, we first consider systems within the class of linear mixed models (LMM) \citep{West07}. This widely applied type of models takes into account the existence of random effects, that is, coefficients of the predictors which are themselves random variables. A functional equation in a linear mixed model has the form:
\begin{equation}
\begin{split}
\mathrm{V}_i = \sum_k \mathrm{b}_{ik} \mathrm{V}_{1k} + \sum_k \epsilon_{ik} \mathrm{V}_{2k} + \xi_i.
\end{split}
\label{elm}
\end{equation}
The sets of parents $\mathbf{V}_1 = \{ \mathrm{V}_{1,1},...,\mathrm{V}_{1,n_1}\}$ and $\mathbf{V}_2 = \{ \mathrm{V}_{2,1},...,\mathrm{V}_{2,n_2}\}$ can overlap. Here $\mathrm{b}_{ik}$ indicates a constant fixed coefficient, while $\epsilon_{ik}$ indicates a random coefficient, that is, $\epsilon_{ik}$ is itself a random variable. For example, $\epsilon_{ik}$ can represent across-subjects variability in the influence strength of a parent variable. All random coefficients are hidden variables. Furthermore, only a subset of $\mathbf{V}$ may be observed. For simplicity, we restrict the examples to Gaussian linear mixed models. Because linear Gaussian models belong to the special family of \cite{Hoyer2009} for which cv-independencies symmetrically hold, this has the advantage that in these examples we can relate cv(nrr)-dependencies only to the presence of random effects introducing nonlinearities in the equations. LMM equations are only in the AN form if the random coefficients vanish. A CAN form can be obtained conditioning on the parents in $\mathbf{V}_2$. We use LMM models for exemplary purpose because the connection between random effects and cv(nrr)-dependencies facilitates the explanation. However, as it is clear from the general form of the functional equations that can have the cv(nrr)-CAN-forms, according to Theorem 1 and 2, cv(nrr)-independencies will exist in a much wider type of systems than LMM models. We will later discuss general versions of these examples, sharing the same causal structures as in Figure \ref{f1}, but with a more general form of the functional equations. Furthermore, note that the random coefficients do not play any especial role other than being hidden variables which appear multiplicatively with the observed variables.

Figure \ref{f1} shows examples of different effects that conditioning has on the cv(or nrr)-independencies asymmetry. For simplification from now on we describe these examples referring only to cv-independencies, but the same reasoning holds for the nrr-independencies. To reflect the form of the equation in the graphical representation, we indicate by an arrow $\mathrm{V}_i \xrightarrow[]{\epsilon} \mathrm{V}_j$ the presence of $\epsilon \mathrm{V}_i$ in the equation of $\mathrm{V}_j$, but as mentioned above the random effects are just hidden variables. We focus on cv-dependencies between $\mathrm{X}$ and $\mathrm{Y}$, conditioned or unconditioned on $\mathrm{Z}$, which are collected in Table \ref{t1}. In Figure \ref{f1}A, conditioning on $\mathrm{Z}$ does not alter the asymmetry. This is because it is the influence of $\epsilon \mathrm{V}$ on $\mathrm{Y}$ what leads to a cv-dependence in the direction $\mathrm{Y} \rightarrow \mathrm{X}$. Because $\mathrm{V}$ is independent of $\mathrm{X}$, $\epsilon \mathrm{V}$ acts effectively as a source of noise on $\mathrm{Y}$ and the equation of $\mathrm{Y}$ has the CAN form for $\mathrm{X}$, conditioned or unconditioned on $Z$. $\epsilon \mathrm{V}$ does not have a Gaussian distribution, which brings the distribution of $\mathrm{X}$ and $\mathrm{Y}$ out of the special family of \cite{Hoyer2009} and leads to cv-dependencies in the direction opposite to causality. In this case, if $\mathrm{V}$ is observable, the collider $\mathrm{V}\rightarrow \mathrm{Y} \leftarrow \mathrm{X}$ can be identified using conditional independencies. Otherwise, $\sigma_{\mathrm{Y}|\mathrm{X}} \perp \mathrm{X}$, $\sigma_{\mathrm{X}|\mathrm{Y}} \notperp \mathrm{Y}$ provides new causal information.

In Figure \ref{f1}B, conditioning on $\mathrm{X}$ activates the collider $\mathrm{V} \rightarrow \mathrm{X} \leftarrow \mathrm{Z}$, activating a path of dependence between $\mathrm{V}$ and $\mathrm{Y}$. Changes in the mean of $\mathrm{V}$ modulate the variance of $\epsilon$, leading to $\sigma_{\mathrm{Y}|\mathrm{X}} \notperp \mathrm{X}$. In the opposite direction, again $\epsilon \mathrm{V}$ acts a source of non-Gaussian noise, leading to $\sigma_{\mathrm{X}|\mathrm{Y}} \notperp \mathrm{Y}$. Conditioning on $\mathrm{Z}$ inactivates the alternative path between $\mathrm{V}$ and $\mathrm{Y}$, providing the CAN form to the equation of $\mathrm{Y}$. The non-Gaussian influence from $\epsilon \mathrm{V}$ results in the asymmetry $\sigma_{\mathrm{Y}|\mathrm{X}, \mathrm{Z}} \perp \mathrm{X}$, $\sigma_{\mathrm{X}|\mathrm{Y}, \mathrm{Z}} \notperp \mathrm{Y}$.

\begin{table}[t]
\caption{Cv-Independencies in the examples of Figure \ref{f1}. The last column indicates whether or not it is possible to infer a potential cause based on the criterion of Proposition $4$.}
\vspace{0.1in}
\label{t1}
\begin{center}
\begin{tabular}{| l | l | l | l | l | l |}
\hline \hline
\  & $\sigma_{\mathrm{Y}|\mathrm{X}}$ & $\sigma_{\mathrm{X}|\mathrm{Y}}$ & $\sigma_{\mathrm{Y}|\mathrm{X}, \mathrm{Z}}$ & $\sigma_{\mathrm{X}|\mathrm{Y}, \mathrm{Z} }$ & \ \\
\hline \hline
A& $\perp$  & $\notperp$  & $\perp$ & $\notperp$ & Yes \\
\hline
B & $\notperp$  & $\notperp$  & $\perp$ & $\notperp$ & Yes\\
\hline
C& $\notperp$  & $\notperp$  & $\perp$ & $\perp$ & No\\
\hline
D& $\perp$  & $\perp$  & $\perp$ & $\notperp$ & No\\
%\hline
%E& $\perp$  & $\perp$  & $\notperp$ & $\notperp$ & No\\
%\hline
%F&$\notperp$  & $\notperp$  & $\notperp$ & $\notperp$ & No \\
\hline
\end{tabular}
\end{center}
\end{table}

In Figure \ref{f1}C, conditioning does not help to find an asymmetry. When $\mathrm{Z}$ is not conditioned, either conditioning on $\mathrm{X}$ or $\mathrm{Y}$ changes the mean of $\mathrm{Z}$, which modulates the variance of $\epsilon$. After conditioning $\mathrm{Z}$, the system is reduced to a linear Gaussian model, leading to a symmetry of cv-independencies. Finally, in Figure \ref{f1}D, conditioning creates a misleading asymmetry. Because the random effect only affects $\mathrm{Z}$, without conditioning $\mathrm{Z}$ the system is linear Gaussian, resulting in a symmetric cv-independence. After conditioning $\mathrm{Z}$, a dependence is created between the random effect and both $\mathrm{X}$ and $\mathrm{Y}$. Because $\mathrm{Y} \perp \mathrm{Z} |\mathrm{X}$, this dependence is inactivated when further conditioning on $\mathrm{X}$, leading to $\sigma_{\mathrm{Y}|\mathrm{X}, \mathrm{Z}} \perp \mathrm{X}$. That is, in this case the cv-independence results from a more general conditional independence. In the opposite direction, $\mathrm{Y}$ cannot inactivate the dependence between $\mathrm{X}$ and $\mathrm{Z}$ ($\mathrm{X} \notperp \mathrm{Z} |\mathrm{Y}$), and the effect of $\epsilon$ leads to $\sigma_{\mathrm{X}|\mathrm{Y}, \mathrm{Z}} \notperp \mathrm{Y}$. Because the asymmetry only appears after conditioning on $\mathrm{Z}$, the extra check of Proposition $4$ can detect that it is not reliable to infer a potential cause from $\mathrm{X}$ to $\mathrm{Y}$.

These examples do not cover all possible effects of conditioning, but indicate that conditioning can maintain an informative asymmetry (Figure \ref{f1}A), create an informative asymmetry (Figure \ref{f1}B), exchange symmetries of cv-dependencies and cv-independencies (Figure \ref{f1}C), and create a misleading asymmetry that has to be detected by the extra checks of Proposition $4$ (Figure \ref{f1}D). Note that the graphs of Figure \ref{f1} do not have the structure of DAGs for all variables, since the random effect variables are assigned to edges instead of nodes. However, the way they provide information about cv(nrr)-independencies suggests that graphical criteria can be used to read cv(nrr)-independencies. A formal introduction of graphical criteria will be described in forthcoming work.

We now discuss more general forms of systems that would lead to the cv-independence asymmetries reported in Table \ref{t1}A-B, corresponding to the causal structures of Figure \ref{f1}A-B, that is, the examples for which it is possible to infer a potential cause from $\mathrm{X}$ to $\mathrm{Y}$. The pattern of cv-independencies of Table \ref{t1}A is more generally compatible with any system of the form:
\begin{equation}
\begin{split}
\mathrm{Z} = \eta_z ; \ \ \mathrm{X} = b_{xz} \mathrm{Z} + \eta_x; \ \  \mathrm{Y} = b_{yx} \mathrm{X} + b_{yz} \mathrm{Z} + f_y(\mathrm{V}, \epsilon, \varepsilon_y),
\end{split}
\label{ee1}
\end{equation}
where $\eta$ indicates a Gaussian noise. We follow the same notational rule as in Section \ref{s4}, writing the functional equations in the most generic form possible given the constraints we require. This class of systems is more general than Gaussian LMM models since $f_y$ can have any form, including nonlinearities, and $\varepsilon_y$ can be non-Gaussian. This is because, with respect to $\mathrm{X}$, the component $f_y(\mathrm{V}, \epsilon, \varepsilon_y)$ acts as an additive noise, in agreement with the CAN form of Eq.\,\ref{e4}. Furthermore, the pattern $\sigma_{\mathrm{Y}|\mathrm{X}, \mathrm{Z}} \perp \mathrm{X}$ and $\sigma_{\mathrm{X}|\mathrm{Y}, \mathrm{Z}} \notperp \mathrm{Y}$ of Table \ref{t1}A, which by itself allows inferring the potential cause from $\mathrm{X}$ to $\mathrm{Y}$, holds for a larger class of systems compatible with the causal structure of Figure \ref{f1}A:
\begin{equation}
\begin{split}
\mathrm{Z} = \varepsilon_z ; \ \ \mathrm{X} = f_x(\mathrm{Z}, \varepsilon_x); \ \  \mathrm{Y} = f_{y, 1}(\mathrm{X},\mathrm{Z}) + f_{y, 2}(\mathrm{Z}, \mathrm{V}, \epsilon, \varepsilon_y),
\end{split}
\label{ee2}
\end{equation}
where all noises can have generic distributions and $f_x$, $f_{y1}$, and $f_{y2}$ are generic and can be nonlinear. Again, after conditioning $\mathrm{Z}$, given the causal structure of Figure \ref{f1}A and the form of the functional equation of $\mathrm{Y}$ in Eq.\,\ref{ee2}, the cv-CAN form holds according to Theorem 1.

In the same way, the pattern of Table \ref{t1}B is also obtained for a much wider class of functional equations compatible with the causal structure of Figure \ref{f1}B:
\begin{equation}
\begin{split}
\mathrm{Z} = \varepsilon_z ; \ \ \mathrm{X} = f_x(\mathrm{Z}, \mathrm{V}, \epsilon, \varepsilon_x);\ \ \mathrm{Y} = f_{y, 1}(\mathrm{X} , \mathrm{Z})+ f_{y, 2}(\mathrm{Z}, \varepsilon_y),
\end{split}
\label{ee3}
\end{equation}
where all noises have generic distributions and $f_x$, $f_{y, 1}$, and $f_{y, 2}$ are generic.

The analysis of these concrete examples illustrates how, when the functional equations are known or hypothesized, Theorem 1 (or Theorem 2), allow determining which cv(or nrr)-independencies exist. In application to data, the criterion of Proposition 4 (or Proposition 5) would be applied after estimating the cv(nrr)-independencies, and the patterns displayed in Table \ref{t1} determine whether a potential cause would be inferred.

\section{Post-nonlinear CAN functions}
\label{s6}

Finally, we also briefly consider how post-nonlinear AN equations \citep{Zhang09} can also be extended to a post-nonlinear CAN form. From Theorem $1$ and $2$, it is straightforward to derive the same conditions for CAN post-nonlinear forms, simply considering that the conditions apply to $h_{i,2}^{-1}(\mathrm{V}_i)$ in Eq.\,\ref{e00}. However, this class of models can be further generalized. To see this, consider a functional equation of the form
\begin{equation}
\label{e5}
\mathrm{Y} = h_4(h_{2}(h_{1}(\mathrm{X},\mathbf{V}_1, \mathbf{U}_1, \varepsilon_y)) + h_3(\mathrm{X}, \mathbf{V}_3, \mathbf{U}_3)),
\end{equation}
where both $h_4$ and $h_2$ are nonlinear invertible functions and $h_1$ is a function that has the CAN form for $\mathrm{X}$ given a certain conditioning set $\mathbf{S}$, where $\mathrm{X}$ is the parent of interest for which we examine the causal relation with $\mathrm{Y}$. The equation can be reexpressed as
\begin{equation}
\label{e6}
h_2^{-1}(h_4^{-1}(\mathrm{Y}) - h_3(\mathrm{X}, \mathbf{V}_3, \mathbf{U}_3)) = h_{1}(\mathrm{X},\mathbf{V}_1, \mathbf{U}_1, \varepsilon_y).
\end{equation}
If $\mathbf{U}_3 = \emptyset$, considering the set $\mathbf{S}'  = \mathbf{S} \cup \mathbf{V}_3$, and using the same notation as in Eq.\,\ref{e4} for the CAN function $h_1$, Eq.\,\ref{e6} has the form
\begin{equation}
\label{e7}
h(\mathrm{Y}, \mathrm{X};\mathbf{S}') = f(\mathrm{X};\mathbf{S}') + \xi_{y|\mathbf{S}'}.
\end{equation}
Exploiting a model of this type requires estimating the functions $h$ and $f$ to minimize the information between $\mathrm{X}$ and $\hat{\xi}_{y|\mathbf{S}'} \equiv \hat{h}(\mathrm{Y}, \mathrm{X};\mathbf{S}')-\hat{f}(\mathrm{X};\mathbf{S}')$. If $\mathrm{X}$ is not an argument of $h_3$ this reduces to the same estimation problem studied in \cite{Zhang09}, with $\hat{\xi}_{y|\mathbf{S}'} \equiv \hat{h}(\mathrm{Y};\mathbf{S}')-\hat{f}(\mathrm{X};\mathbf{S}')$.

The form of Eq.\,\ref{e5} suggests a generalization by an iterative composition of two operations. Consider the operation consisting in an invertible nonlinear univariate transformation $g(z)$ and the operation consisting in the bivariate sum $s(z_1,z_2) = z_1+z_2$. Starting from a function $h(\mathrm{X},\mathbf{V}, \mathbf{U}, \varepsilon_y)$ that has the CAN form for $\mathrm{X}$ given a certain conditioning set $\mathbf{S}$, a set of invertible nonlinear functions $g_i(z)\ i= 1,...,m$ and a set of arguments $z_{2,i} = \tilde{f}_i(\mathrm{X}, \mathbf{V}_i),\ i = 1,...,m$, the functional equation of $\mathrm{Y}$ can be constructed by the iterative composition starting as $g_2(g_1(h)+z_{2,1})+z_{2,2}$, with $s_k(g_k(s_{(k-1)}(g_{(k-1)}, z_{2,k-1})), z_{2,k})$. Because all functions $g_k$ are invertible, the functional equation of $\mathrm{Y}$ can be expressed in the form of Eq.\,\ref{e7} by inverting the operations. As in the case of Eq.\,\ref{e5}, if $\mathrm{X}$ is not an argument of the functions $\tilde{f}_k$, the expression further simplifies to the form studied in \cite{Zhang09}. The required conditioning set is $\bigcup \{\mathbf{S}, \mathbf{V}_1,..., \mathbf{V}_m\}$. The same procedure can be followed replacing the sum operation by the product. This procedure results in increasingly complex functional equations for which in principle cv-independencies and nrr-independencies can be tested. In practice, the difficulty of the estimation problem of Eq.\,\ref{e7} will depend on the number of these operations, the extra variables introduced in the functions analogous to $h_3$, as well as on the number of variables in $h_1(\mathrm{X},\mathbf{V}_1, \mathbf{U}_1, \varepsilon_y)$, and the complexity of the functions.

\section{Conclusions}

In this paper we extended the theory behind the AN framework for structure learning in several ways. We first introduced an alternative regression-free test of independence. This test does not require the reconstruction of the additive noise using the residuals of a nonlinear regression. Instead of testing the independence between the residuals and the parents of a variable (nrr-independencies), it evaluates indirectly the independence between the noise variance and the parents using conditional variances (cv-independencies). The use of cv-independencies is expected to be especially useful when the form of the functional equation is complex. In that case, the family of regression models used may not be powerful enough to capture the form of the actual dependencies, and thus our indirect estimate of independencies may be particularly beneficial. On the other hand, the examination of cv-independencies and nrr-independencies is not mutually exclusive and could be combined to improve learning.

We formulated all the other contributions of this work both for cv-independencies as well as for nrr-independencies. In the latter case, the implementation of nonlinear regressions developed in previous work \citep[see the actual implementations provided by][]{Hoyer2009, Mooij09, Peters2014, Buhlmann14} can already be applied to implement this extended framework. We generalized AN models to partial conditionally-additive-noise (CAN) models with hidden variables. In these models, only some functional equations and only for certain parents have the AN form, possibly after conditioning. We determined when a functional equation has the CAN form that results in cv(or nrr)-independencies. Exploiting asymmetries in cv(or nrr)-independencies, we then introduced a criterion to infer the causal relation between specific pairs of variables in a multivariate system with hidden variables, without restrictions on the form of the functional equations. The criterion can be applied locally, if the CAN form holds for a certain functional equation, and without inferring a global causal ordering \citep{Mooij09}. Because the type of functional equations that have a CAN form is substantially larger than the type of pure additive-noise functional equations, we can expect that cv(nrr)-independencies induced by the CAN form will exist more often and hence that in more practical cases the AN framework will increase the inferential power of standard methods based on conditional independencies. The magnitude of this increase will be specific to each domain of application, depending on the properties of the generative functional equations.

The new criterion can readily be applied to complement the existing algorithms that in the presence of hidden variables extract equivalence classes of causal structures given conditional independencies \citep{Spirtes00, Drton17, Heinze17}. Like for any standard rule of causal orientation used in constraint-based structure learning algorithms \citep[e.g.\,][]{Spirtes00}, this new criterion relies on faithfulness assumptions. While it is an ongoing subject of research to understand when faithfulness holds \citep{Uhler13}, only under these types of assumptions the corresponding analysis of independencies can be applied for structure learning. In future work we will address in full detail how to exploit the new criterion in combination with conditional independencies as part of a structure learning algorithm.

\section*{Acknowledgments}

This research was supported by the NIH Brain Initiative (Grant No. U19 NS107464) and by the Fondation Bertarelli.

\section*{Appendix}

\subsection*{Proof of Proposition $3$}

\noindent Proof of Proposition $3$: We first prove that cv-independence faithfulness implies nrr-independence faithfulness. Consider that a nonlinear regression is implemented such that $ \hat{\varepsilon}_y \equiv \mathrm{Y}-\hat{f}_y(\mathrm{X},\mathbf{S})$ is independent of $\mathrm{X}$ despite $\mathrm{Y} \in \mathbf{Pa}_x$. Then the statistical model $\mathrm{Y} = \hat{f}_y(\mathrm{X},\mathbf{S}) + \hat{\varepsilon}_y$ has the AN form and it follows that $\sigma_{\mathrm{Y}|\mathrm{X},\mathbf{S}} \perp \mathrm{X}$. Given that $\hat{\varepsilon}_y \perp \mathrm{X}$ implies $\sigma_{\mathrm{Y}|\mathrm{X},\mathbf{S}} \perp \mathrm{X}$, inversely $\sigma_{\mathrm{Y}|\mathrm{X},\mathbf{S}} \notperp \mathrm{X}$ implies $\hat{\varepsilon}_y \notperp \mathrm{X}$. Because cv-independence faithfulness assumes $\sigma_{\mathrm{Y}|\mathrm{X},\mathbf{S}} \notperp \mathrm{X}\ \forall \mathbf{S}\ \subseteq \mathbf{ND}(\mathrm{X}), \mathbf{\mathbf{Pa}}_x \backslash \mathrm{Y} \subseteq \mathbf{S}$, this implies $\hat{\varepsilon}_y \notperp \mathrm{X}\ \forall \mathbf{S}\ \subseteq \mathbf{ND}(\mathrm{X}), \mathbf{\mathbf{Pa}}_x \backslash \mathrm{Y} \subseteq \mathbf{S}$, which corresponds to the assumption of nrr-independence faithfulness. We now justify that nrr-independence faithfulness does not imply cv-independence faithfulness. To see this, it suffices to realize that nrr-independence requires that all moments of the residuals variable $\hat{\varepsilon}_y$ are independent of $\mathrm{X}$. On the other hand, cv-independence only requires that the variance of the residuals variable is independent. The distribution $p(\hat{\varepsilon}_y|x, s)$ can be such that the dependence only appears in the third and higher moments. In that case, cv-independence holds despite nrr-dependence. $\ \ \Box$ %However, because faithfulness regards conditional distributions in the non-generative direction, these type of distributions in which dependencies appear only in higher-order moments can be expected to be not only rare but also fragile to changes in the distribution of the causes. This means that in practice cv-independence faithfulness is not substantially more restrictive.

\subsection*{Proof of Theorem $1$ and Theorem $2$}

We first prove the if and only if conditions of Theorem 1 for the functional equation of $\mathrm{Y}$ to be in the cv-CAN form with respect to a parent $\mathrm{X}$ given the set $\mathbf{S}$ when $\mathrm{X}$ is adjacent to all other potential causes of $\mathrm{Y}$.

\noindent Proof of Theorem $1$: We proceed justifying the necessary and sufficient requirements for each set of hidden and observed variables of Eq.\,\ref{e2}. First, we need $\mathbf{U}_{1,1} = \emptyset$ because $\mathrm{X}$ modulates the variance of any $\mathrm{U} \in \mathbf{U}_{1,1}$, since they appear together as arguments of $f_{1,1}$ in Eq.\,\ref{e2}, which is nonlinear. Also for $\mathrm{U} \in \mathbf{U}_{1,2}$, dependencies on $\mathrm{X}$ produce a change in the variance of $f_{1,2}(\mathbf{V}_{1,2}, \mathbf{U}_{1,2})$ due to nonlinearities, even if $\mathbf{V}_{1,2}$ is conditioned. Because $\mathbf{U}_{1,2}$ are hidden, we have to require $\mathrm{X} \perp \mathrm{U}_k|\mathbf{S}\ \forall \mathrm{U}_k \in \mathbf{U}_{1,2}$. For the same reason, we need $\mathrm{X} \perp \mathrm{U}_k|\mathbf{S}\ \forall \mathrm{U}_k \in \mathbf{U}_{2}$. On the other hand, the variables $\mathbf{U}_{1,4}$, conditioning on $\mathbf{\tilde{V}}_{1,4}$, contribute linearly to $\mathrm{Y}$, similarly to $\mathbf{U}_{3}$. Accordingly, to avoid that these terms introduce a dependence of $\sigma_{\mathrm{Y}|\mathrm{X},\mathbf{S}}$ on $\mathrm{X}$, it is required that $\sigma_{\mathrm{U}_k|\mathrm{X},\mathbf{S}} \perp \mathrm{X} \ \forall \mathrm{U}_k \in \{\mathbf{U}_{1,4} , \mathbf{U}_{3}\} $ and that the covariances fulfill $\sigma_{\mathrm{Z}_i\mathrm{Z}_j|\mathrm{X},\mathbf{S}} \perp \mathrm{X}\ \forall \mathrm{Z}_i, \mathrm{Z}_j \in \mathbf{S}_2$, where $\mathbf{S}_2 = \{ \mathrm{Y}_{1,2}, \mathrm{Y}_2 , \mathbf{U}_{1,4} , \mathbf{U}_{3} \}$.

Regarding the observable variables, we need $\mathbf{V}_{1,1} \subseteq \mathbf{S}$ because of the nonlinearity of $f_{1,1}$ such that $\mathrm{X}$ modulates their variance. Similarly, we also need $\mathbf{V}_{1,2} \subseteq \mathbf{S}$ because of the nonlinearity of $f_{1,2}$ and also because $\mathrm{V} \in \mathbf{V}_{1,2}$ can modulate the variance of variables in $\mathbf{U}_{1,2}$. The same holds to require $\mathbf{\tilde{V}}_{1,3} \subseteq \mathbf{S}$ and $\mathbf{\tilde{V}}_{1,4} \subseteq \mathbf{S}$. We also need $\mathbf{V}_{2} \subseteq \mathbf{S}$ because any $\mathrm{V} \in \mathbf{V}_{2}$ modulates the variance of $\varepsilon_y$. Regarding $\mathbf{V}_{3}$, because these variables only contribute linearly, we can divide the set in two groups. The variables in $\mathbf{\bar{V}}_{3} \equiv \{\mathrm{V}_i \in \mathbf{V}_{3}: \ \sigma_{V_i|\mathrm{X}, \mathbf{S}} \perp \mathrm{X} \}$ may not need to be conditioned, because although not independent from $\mathrm{X}$, the dependence does not affect their conditional variance. Similarly, once we have conditioned on $\mathbf{\tilde{V}}_{1,3}$, the variables $\mathbf{V}_{1,3}$ contribute linearly to $\mathrm{Y}$, and hence can be divided analogously to $\mathbf{V}_{3}$. The variables in $\mathbf{\bar{V}}_{1,3} \equiv \{ \mathrm{V}_i \in \mathbf{V}_{1,3}: \sigma_{V_i|\mathrm{X}, \mathbf{S}} \perp \mathrm{X} \}$ may not need to be conditioned. Because $\mathbf{V}_{3}$ and $\mathbf{V}_{1,3}$ are observable there is the option to try to find a valid $\mathbf{S}$ conditioning on all of them. Alternatively, we can exclude from the conditioning set some subsets $\mathbf{V}_{3,1} \subseteq \mathbf{\bar{V}}_{3}$ and $\mathbf{V}_{1,3,1} \subseteq \mathbf{\bar{V}}_{1,3}$. We complementarily define $\mathbf{V}_{3,2} = \mathbf{V}_{3} \backslash \mathbf{V}_{3,1}$ and $\mathbf{V}_{1,3,2} = \mathbf{V}_{1,3} \backslash \mathbf{V}_{1,3,1}$. If $\mathbf{V}_{3,1} \neq \emptyset$ or $\mathbf{V}_{1,3,1}\neq \emptyset$, we also need to require that the covariance between any pair of variables from these subsets is not modulated by $\mathrm{X}$. Similarly we need that the covariance with the other linear terms in $\mathbf{S}_2$ is also independent of $\mathrm{X}$. This together guarantees that the linear contributions of the functional equation do not create a dependence of $\sigma_{\mathrm{Y}|\mathrm{X},\mathbf{S}}$ on $\mathrm{X}$. Altogether, the observable variables to be included in $\mathbf{S}$ are $\{ \mathbf{V}_{1,1} , \mathbf{V}_{1,2}, \mathbf{\tilde{V}}_{1,3}, \mathbf{V}_{1,3,2}, \mathbf{\tilde{V}}_{1,4}, \mathbf{V}_{2}, \mathbf{V}_{3,2} \}$. Because for each set of variables we described the requirements necessary and sufficient to eliminate their contribution to any dependence of the conditional variance, the fulfillment of these requirements leads to cv-independence. $ \Box$

\noindent Proof of Theorem $2$: The proof is analogous to the one of Theorem 1. We only highlight the differences. In contrast to cv-independence, nrr-independence regards all moments of the residuals. According to Eq.\,\ref{e3}, the variables $\mathrm{Z} \in \{\mathbf{V}_{1,3,1}, \mathbf{V}_{3,1}, \mathbf{U}_{1,4}, \mathbf{U}_{3}\}$ contribute linearly to $\mathrm{Y}$ after conditioning on $\mathbf{S}$. The regression will result in a residual for $\mathrm{Y}$ that can equally be decomposed as comprising a contribution from each of these linearly additive terms, which is proportional to the residual from the separate regression of each $\mathrm{Z}$. That is, the regression $\hat{f}_y(\mathrm{X};\mathbf{S})$ will contain a component fitted to each conditional mean $\mu_{z|\mathrm{X};\mathbf{S}}$. Accordingly, the residual of $\mathrm{Y}$ has a contribution from the residuals $\varepsilon_{z|\mathrm{X};\mathbf{S}} \equiv \mathrm{Z}-\mu_{z|\mathrm{X};\mathbf{S}}$, and hence any dependence of $Z$ with $\mathrm{X}$ in any moment other than the mean produces an nrr-dependence between $\mathrm{X}$ and the residual of $\mathrm{Y}$. $\ \Box$

\subsection*{The special family of distributions with bidirectional statistical CAN form}

We here show that, analogously to the case of pure AN equations studied in \cite{Hoyer2009}, there is a special family of joint distributions $p(\mathrm{X},\mathrm{Y}|\mathbf{S})$ that allows a CAN statistical form in both directions. The proof relies on the one of \cite{Hoyer2009}. It suffices to realize that when the functional equation of $\mathrm{Y}$ admits the CAN form for $\mathrm{X}$ given the set $\mathbf{S}$, the bivariate distribution $p(\mathrm{X},\mathrm{Y}|\mathbf{S})$, for fixed $\mathbf{S}$, can be expressed in the same form used in the proof of \cite{Hoyer2009}. In particular, using the notation of Eq.\,\ref{e4},
\begin{equation}
\label{ss0}
\log p(\mathrm{X},\mathrm{Y}|\mathbf{S}) = \log p_{\xi_{y|\mathbf{S}}}(\mathrm{Y}-f_{1,1}(\mathrm{X} ;\mathbf{S}))+ \log p(\mathrm{X}),
\end{equation}
which is analogous to Equation 5 in \cite{Hoyer2009}. The rest of the proof follows equivalently.

%\bibliography{spikedistancesyPRE3}

\end{document}